\documentclass[conference]{IEEEtran}
\IEEEoverridecommandlockouts
\usepackage{cite}
\usepackage{microtype}[kerning=True]
\usepackage{amsmath,amssymb,amsfonts}
\usepackage{algorithmic}
\usepackage{graphicx}
\usepackage{textcomp}
\usepackage{xcolor}
\def\BibTeX{{\rm B\kern-.05em{\sc i\kern-.025em b}\kern-.08em
    T\kern-.1667em\lower.7ex\hbox{E}\kern-.125emX}}
\begin{document}

\title{Multimodal Anomaly Detection for Human-Robot Interaction\\

\thanks{This work was supported by the HORIZON Europe Programme - HARIA Project, Grant agreement 101070292 and by FCT through the LASIGE Research Unit, ref. UID/00408/2023} }
\author{\IEEEauthorblockN{ Guilherme Ribeiro}
\IEEEauthorblockA{\textit{LASIGE}\\\textit{Faculty of Sciences}\\
U. Lisboa, Portugal \\
grribeiro@lasige.di.fc.ul.pt}
\and
\IEEEauthorblockN{  Iordanis Antypas}
\IEEEauthorblockA{\textit{LASIGE} \\
\textit{Faculty of Sciences}\\
U. Lisboa, Portugal }
\and
\IEEEauthorblockN{ Leonardo Bizzaro}
\IEEEauthorblockA{\textit{Dep. of Information Engineering} \\
\textit{University of Padova}\\
Padova, Italy }
\and
\IEEEauthorblockN{ João Bimbo}
\IEEEauthorblockA{\textit{LASIGE} \\
\textit{Faculty of Sciences}\\
U. Lisboa, Portugal }
\and
\IEEEauthorblockN{ Nuno Cruz Garcia}
\IEEEauthorblockA{\textit{LASIGE} \\
\textit{Faculty of Sciences}\\
U. Lisboa, Portugal }
}

\maketitle
\vspace*{-20pt}
\begin{abstract}
Ensuring safety and reliability in human-robot interaction (HRI) requires the timely detection of unexpected events that could lead to system failures or unsafe behaviours. Anomaly detection thus plays a critical role in enabling robots to recognize and respond to deviations from normal operation during collaborative tasks. While reconstruction models have been actively explored in HRI, approaches that operate directly on feature vectors remain largely unexplored. In this work, we propose MADRI, a framework that first transforms video streams into semantically meaningful feature vectors before performing reconstruction-based anomaly detection. Additionally, we augment these visual feature vectors with the robot’s internal sensors' readings and a Scene Graph, enabling the model to capture both external anomalies in the visual environment and internal failures within the robot itself. To evaluate our approach, we collected a custom dataset consisting of a simple pick-and-place robotic task under normal and anomalous conditions. Experimental results demonstrate that reconstruction on vision-based feature vectors alone is effective for detecting anomalies, while incorporating other modalities further improves detection performance, highlighting the benefits of multimodal feature reconstruction for robust anomaly detection in human-robot collaboration.
\end{abstract}


\section{Introduction}
The increasing integration of robotic systems into human-centric environments, such as logistics, manufacturing, and healthcare, underscores the critical importance of Human-Robot Interaction (HRI) \cite{goodrich2008human}. This coexistence offers significant benefits, but it also introduces a fundamental challenge: ensuring the safety and reliability of these collaborative systems. While standard safety mechanisms and established protocols are often implemented to mitigate known risks, they are inherently limited in their ability to address anomalous behaviours, behaviours that do not conform to a well-defined notion of normal\cite{goodrich2008human}. The ability of a robot to autonomously detect and respond to these unforeseen anomalies is thus a crucial step toward achieving truly robust and safe human-robot collaboration.

Anomaly detection in robotics typically involves identifying data points or patterns that deviate significantly from a defined ``normal" state \cite{chandola2009anomaly}. Given that anomalous events are rare and difficult to capture in sufficient quantities for supervised training, unsupervised learning techniques are a natural fit for this problem. A widely adopted and successful approach involves the use of reconstruction models, such as autoencoders, which are trained on a large dataset of normal behaviour to learn a compact representation of the data \cite{gong2019memorizing}. During inference, a high reconstruction error for a given input signals a potential anomaly. However, many of these methods operate on raw input data, such as pixel values from video streams, which can be computationally expensive and struggle to generalize under dynamic environmental conditions. A promising, yet fairly unexplored, alternative is to perform anomaly detection in a semantically meaningful feature space, where the model can focus on the high-level attributes of the scene rather than pixel-level details.
\begin{figure}
    \centering
    \includegraphics[width=0.49\linewidth]{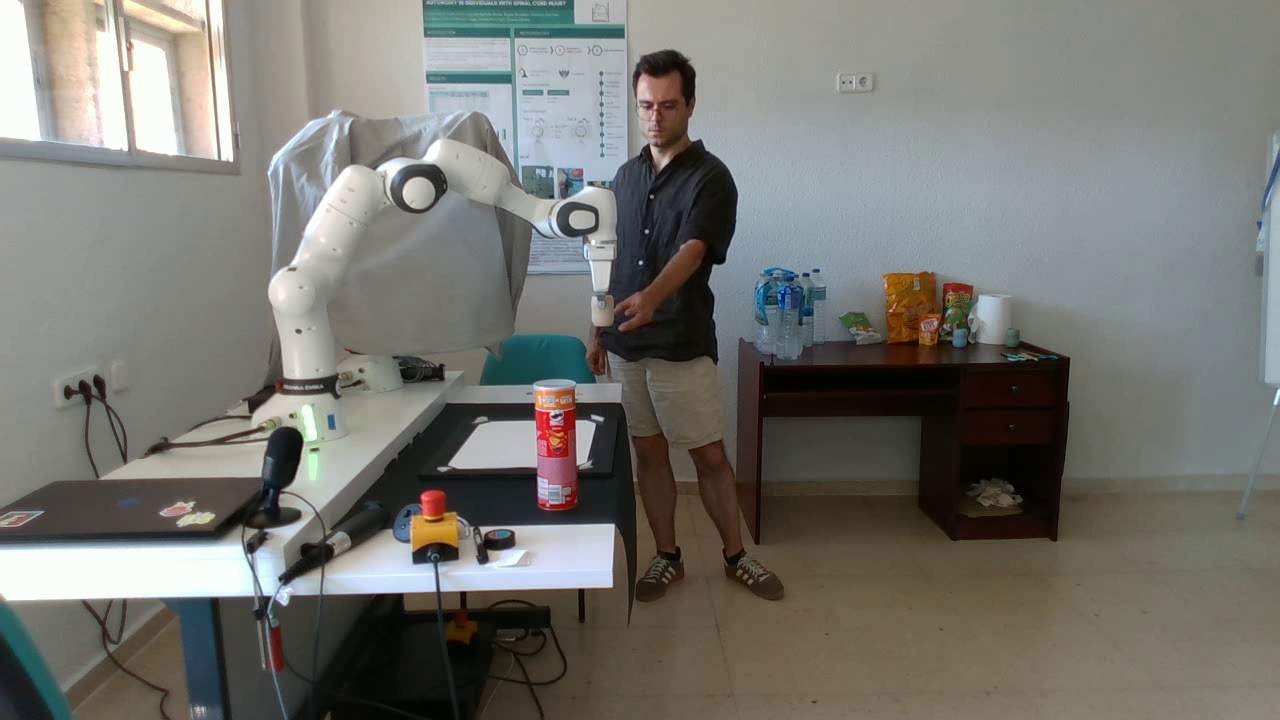}
    \includegraphics[width=0.49\linewidth]{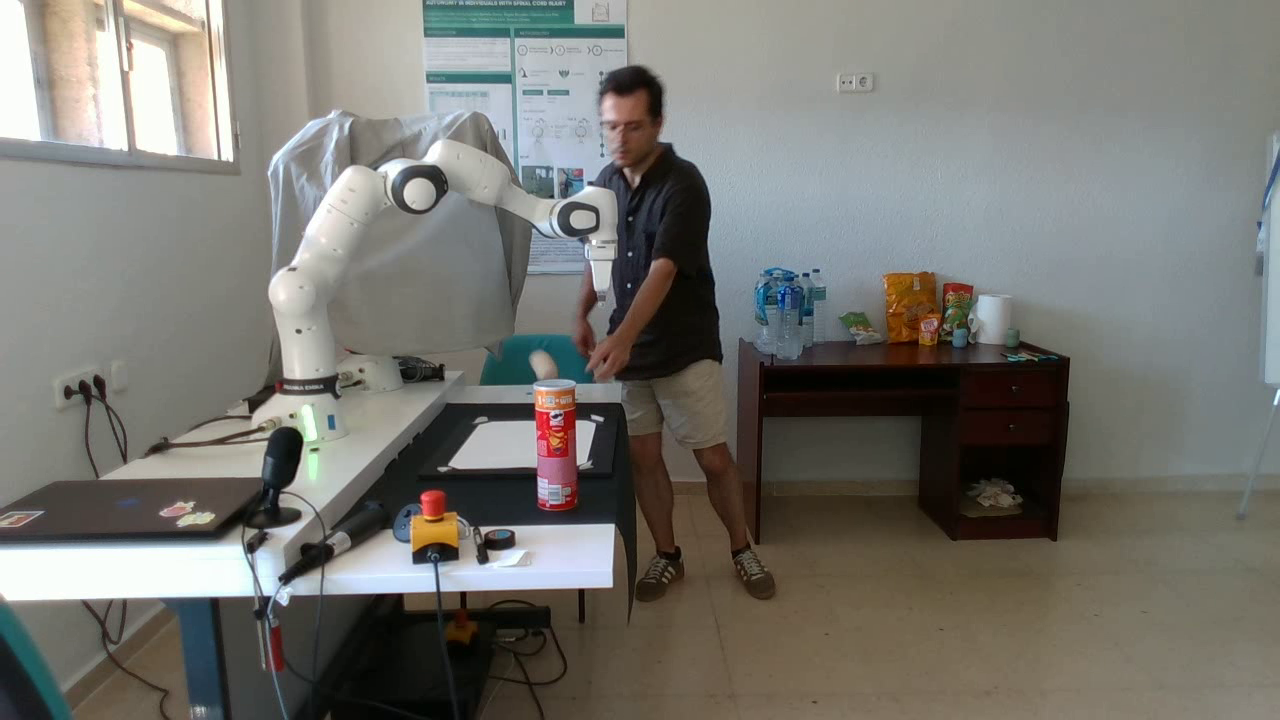}
    \caption{Human Robot Cup Handover Failure. The human did not position the cup exactly as expected, resulting in a grasp failure and the robot dropping the cup.}
    \label{fig:Failure Example}
\end{figure}

In this work, we propose a novel framework for anomaly detection in HRI that addresses these limitations. We introduce MADRI (Multimodal Anomaly Detection for Human-Robot Interaction), a framework that leverages reconstruction on multimodal data to enhance anomaly detection performance. Our approach first transforms video streams into semantically rich feature vectors. We then augment these visual features with data from the robot's internal state, specifically joint torque sensor readings, and from a Scene Graph which allows the model to capture both external environmental deviations and internal mechanical failures. This multimodal strategy provides a comprehensive view of the system's state, enabling more robust anomaly detection. To validate our approach, we collected a custom dataset of a pick-and-place robotic task under various normal and anomalous conditions, providing a practical evaluation platform for our method. We further compared MADRI against a full video reconstruction baseline to demonstrate the benefits of incorporating multimodal information.

The specific contributions of this paper are as follows:

\begin{itemize}
\item We propose MADRI, a multimodal anomaly detection framework that effectively combines video streams, robot sensory data and scene graph to accurately identify anomalies in HRI tasks.
\item We demonstrate that performing reconstruction-based anomaly detection on a semantically meaningful feature space is an effective strategy for video-based anomaly detection.
\item We present a custom human-robot handover dataset specifically designed for evaluating multimodal anomaly detection systems.
\item We conduct a thorough experimental evaluation, demonstrating that incorporating different modalities significantly improves anomaly detection performance over vision-only methods, thereby validating the benefits of a multimodal approach for robust HRI.
\end{itemize}

\section{Methodology}
\begin{figure*}[!h]
    \centering
    \includegraphics[width=0.7\linewidth]{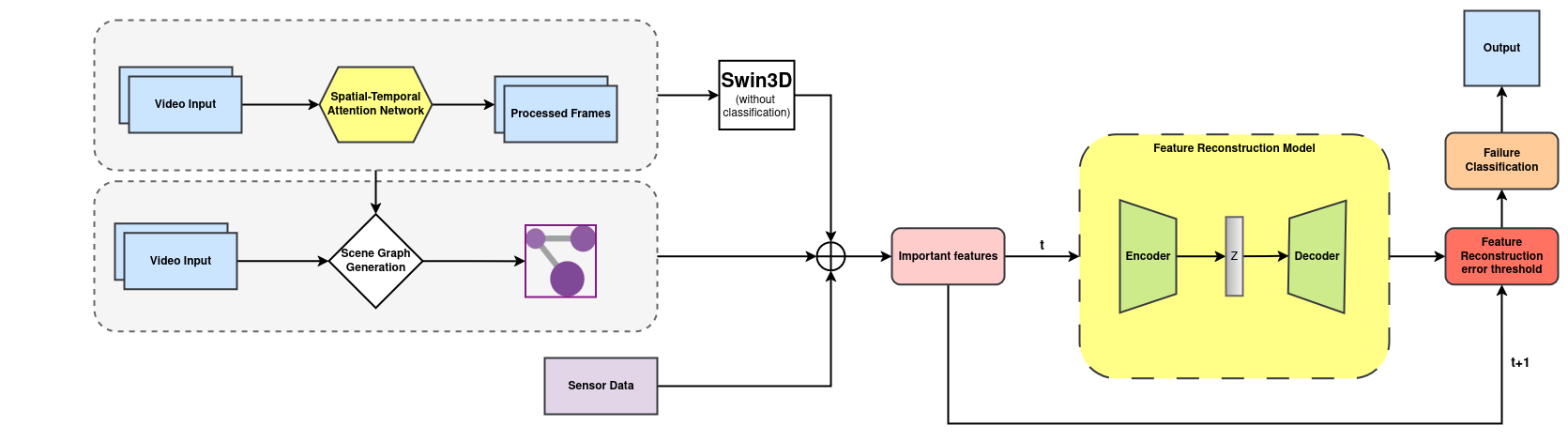}
    \caption{Overview of the proposed failure detection framework. Video input is processed by a Swin3D backbone (without classification head) to extract visual features, which can optionally be fused with sensor data. The resulting feature vector is passed through a feature reconstruction model composed of an encoder-decoder architecture. Reconstruction error is then compared against a predefined threshold to classify failure and generate the final output}
    \label{fig:overall framework}
\end{figure*}

\subsection{Overview}
Our MADRI framework leverages RGB video data, robot joint sensor readings and Scene Graph to perform feature-level anomaly detection in human-robot interaction scenarios. We propose a multimodal anomaly detection approach that is capable of identifying anomalies across three distinct dimensions: visual behaviour (captured through RGB video), physical dynamics (captured through robot’s internal sensors) and relational (captured through the Scene Graph). This multi-modality enables the system to detect not only visually apparent anomalies but also subtler, sensor-level deviations that may not be visible or relationships between people/objects that aren't correct.

To extract meaningful features from the video, we fine-tune a pre-trained Swin3D model on a human-robot interaction action recognition task. After fine-tuning, we remove the final classification layer, allowing the model to output high-level semantic feature vectors instead of class probabilities. These vectors represent the core visual understanding of the action in each clip.

We augment these visual features by concatenating them with torque values from the robot’s joints and a matrix representing the Scene Graph, creating a richer, multimodal feature representation. This combined feature vector is then passed to a Feature Reconstruction Model, an autoencoder trained to learn and reconstruct only the distribution of normal behaviour.

During inference, we compute the reconstruction error for each feature vector. A threshold is defined and any reconstruction error exceeding this threshold is flagged as anomalous. The complete pipeline is illustrated in Fig.~\ref{fig:overall framework}.

\subsection{Visual Feature Extraction}
To extract visual features efficiently and effectively, we fine-tuned the pre-trained Swin3D\cite{liu2021videoswintransformer} model available through PyTorch \cite{paszke2019pytorchimperativestylehighperformance}. This model was originally trained on the Kinetics dataset, a large-scale benchmark for human action recognition \cite{carreira2017quo}. Given the conceptual similarity between human-object interaction and human-robot interaction, leveraging a model pre-trained on Kinetics provides a strong foundation for recognizing human-robot interaction behaviours.

\begin{figure}[!h]
    \centering
    \includegraphics[width=0.8\linewidth]{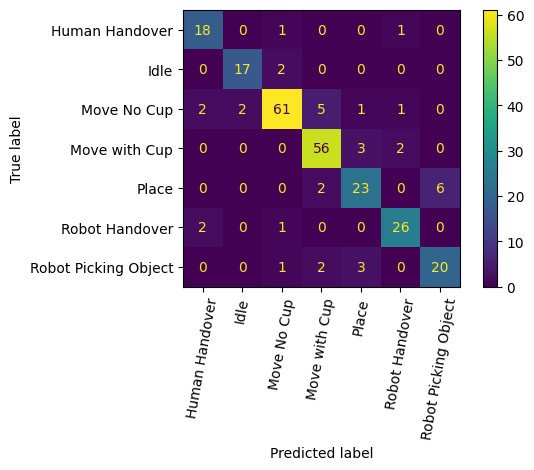}
    \caption{Confusion matrix for the action recognition model across seven action classes. The diagonal values indicate correct classifications, while off-diagonal values represent misclassification. The model demonstrates high accuracy for classes such as 'Move No Cup' and 'Move with Cup', with some confusion between visually similar actions like 'Place' and 'Robot Picking Object'.}
    \label{fig:Action Recognition CM}
\end{figure}

For the fine-tuning stage, we defined the following action classes: \textit{Idle, (Robotic) Move with no Cup, Human Handover, (Robotic) Move with Cup, (Robotic) Place, Robot Picking Object,} and \textit{Robot Handover}. These categories capture the essential dynamics involved in the interaction tasks. The resulting performance is summarized in the confusion matrix shown in Fig. \ref{fig:Action Recognition CM}.

Once fine-tuning was complete, we removed the final classification layer of the Swin3D model, yielding a 768-dimensional feature vector for each input clip. This transformation serves a critical purpose: directly reconstructing full video frames or images during anomaly detection is computationally intensive and often includes reconstruction of irrelevant visual features (e.g., background textures, ceiling, or lighting conditions) that do not contribute meaningfully to the task. By working in the feature space instead, we not only accelerate the reconstruction process but also focus on semantically relevant components of the scene, such as object manipulation and motion patterns. These feature vectors are the inputs used for the vision-only anomaly detection module.


\subsection{Sensor Data Integration}

During task execution, the robot continuously streams sensor data, including joint torques and gripper positions, at a high frequency. To manage the resulting high-dimensional sensor matrix and reduce computational overhead, we applied Max Pooling across the temporal axis. This allowed us to downsample the data while preserving the most salient and informative sensor readings. The resulting pooled sensor features were then concatenated with the corresponding visual feature vectors.

\subsection{Scene Graph}

We employ the pre-trained OED model provided by the authors \cite{wang2024oedonestageendtoenddynamic} to construct a temporal Scene Graph for each clip. In this graph, the nodes represent humans and objects, while the edges capture the relationships between them. To align this representation with our feature structure, we generate a matrix in which each possible detected object is evaluated against the set of relations present in the clip. Finally, this matrix is concatenated into the overall feature vector.

\subsection{Reconstruction Model}

The reconstruction model is implemented as a straightforward autoencoder architecture. It compresses the input features into a lower-dimensional latent space and then reconstructs them back to their original shape. Each layer in the encoder and decoder consists of a standard linear block, structured as follows: (1) a Linear layer, (2) Batch Normalization, (3) ReLU activation, and (4) Dropout for regularization. The model is trained using L1 Loss.

The depth and capacity of the model are adjusted based on the specific requirements of each task. For vision-only reconstruction, a shallower architecture suffices due to the lower complexity of the input. In contrast, multimodal reconstruction—
, which involves integrating visual and sensor data, requires a deeper network to adequately capture and reconstruct the richer joint representation. 

\subsection{Anomaly Scoring}

After obtaining the reconstruction errors for all frames in a video, we estimate what constitutes normal activity within that specific sequence. To do this, we first normalize the reconstruction errors between zero and one. Then, for all possible thresholds, we calculate the F1-Score. The threshold that presented the best score, was the chosen one. In the error reconstruction vector, any value above the threshold is considered to be an anomaly. 

\section{Experimental Setup}

To evaluate the proposed method, we proceeded to collect our own domain-specific dataset. This custom dataset enabled us to test our framework in a more relevant and controlled environment. Additionally, we carried out an ablation study to assess the contribution of each component within our approach. This section also outlines the evaluation metrics used throughout our experiments.

\subsection{Implementation Details}

Since we use a pre-trained model, it is essential to preprocess our data in the same way as the original training data to ensure compatibility. Each video undergoes the following preprocessing steps: (1) The frame rate is standardized to 15 FPS; (2) Videos are segmented into clips of 32 consecutive frames; (3) Each frame is resized to 256 pixels on the shorter side; (4) Bilinear interpolation is applied to maintain visual quality; (5) A central crop of size 224×224 is extracted from each frame; (6) Pixel values are rescaled to the [0, 1] range; (7) Finally, normalization is applied using a mean of [0.485,0.456,0.406] and a standard deviation of [0.229,0.224,0.225], matching the preprocessing used during the model’s original training.


This means that anomaly detection is performed at the clip level rather than the full video level, enabling the system to pinpoint the precise moment an anomaly occurs.

\subsection{Dataset Collection}

\begin{figure}
    \centering
    \includegraphics[width=0.49\linewidth]{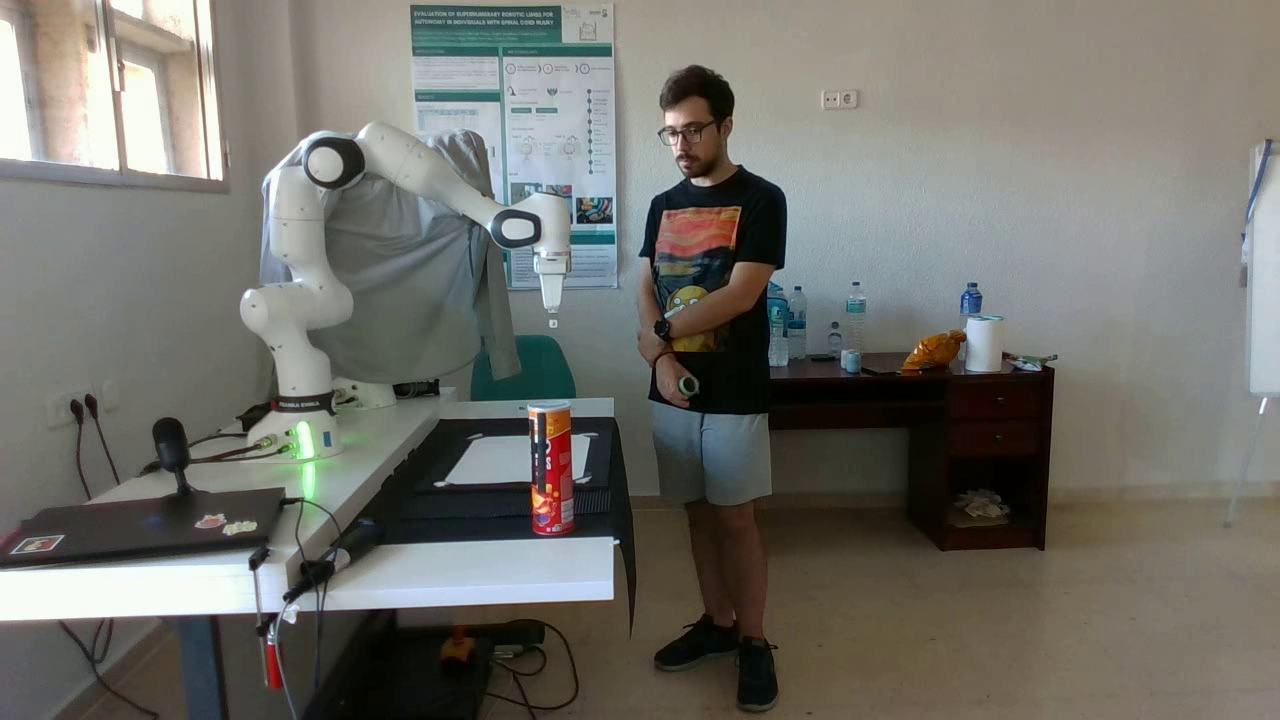}
    \includegraphics[width=0.49\linewidth]{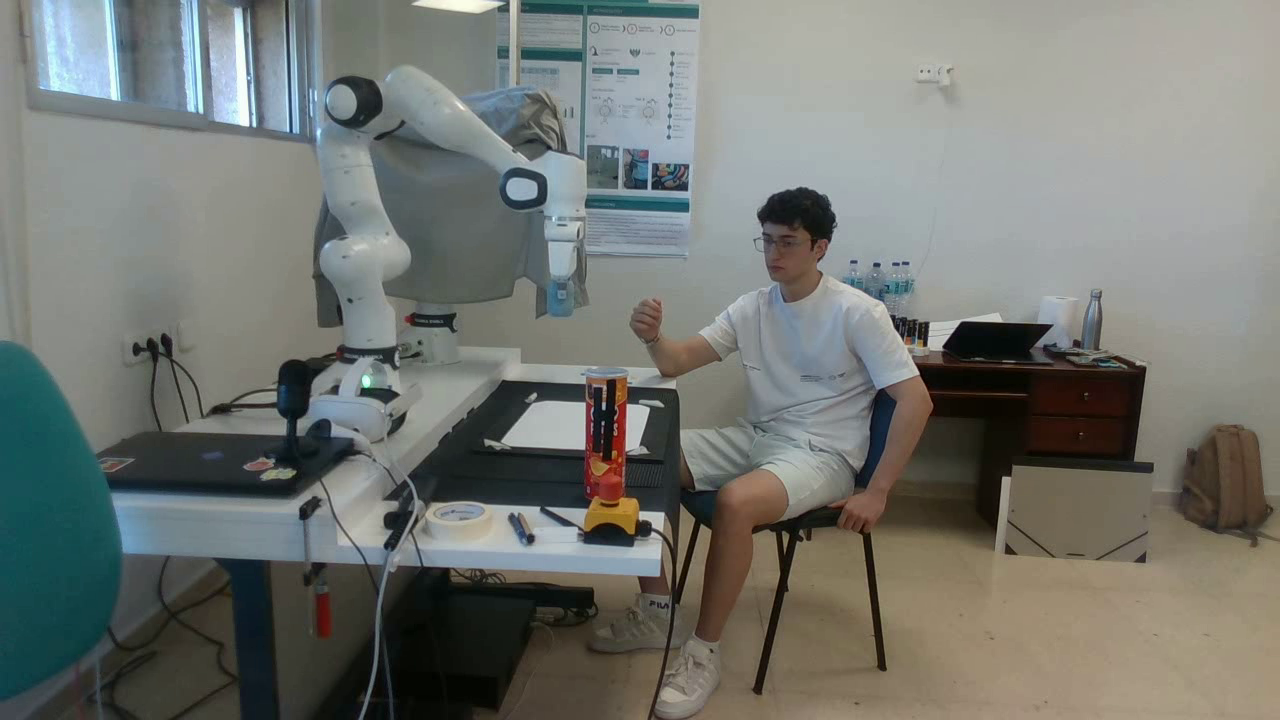}
    \includegraphics[width=0.49\linewidth]{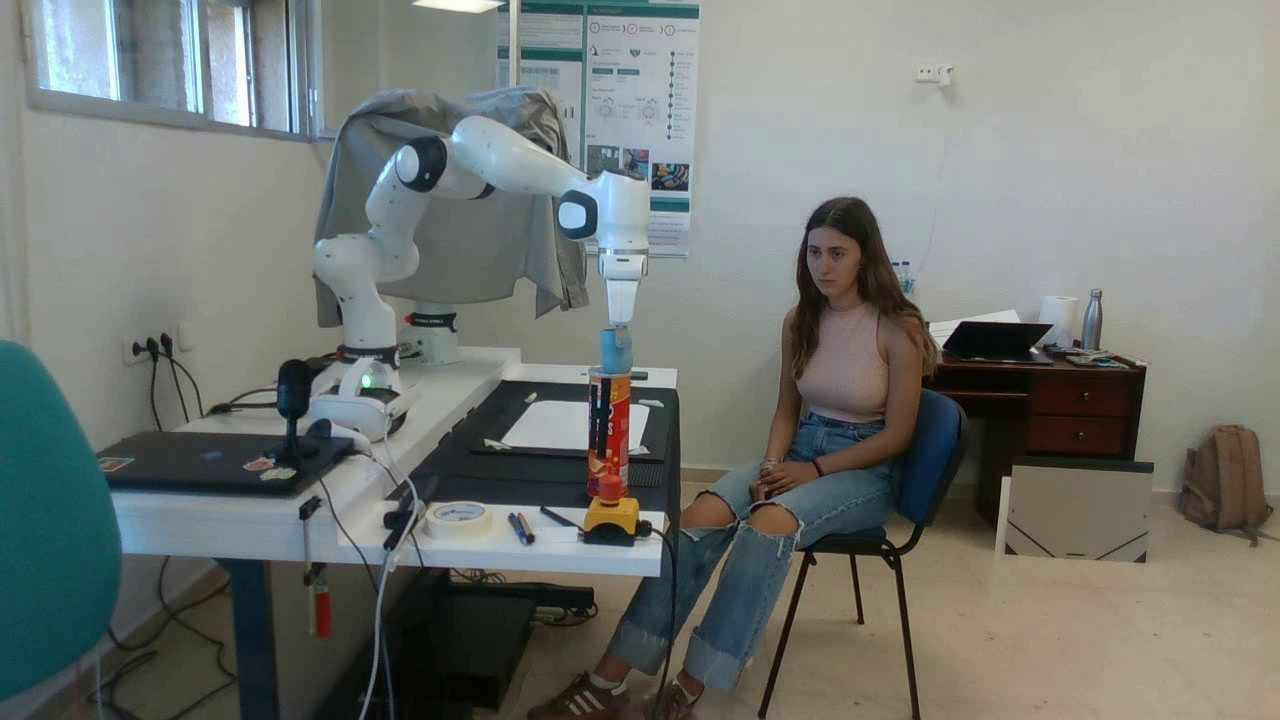}
    \includegraphics[width=0.49\linewidth]{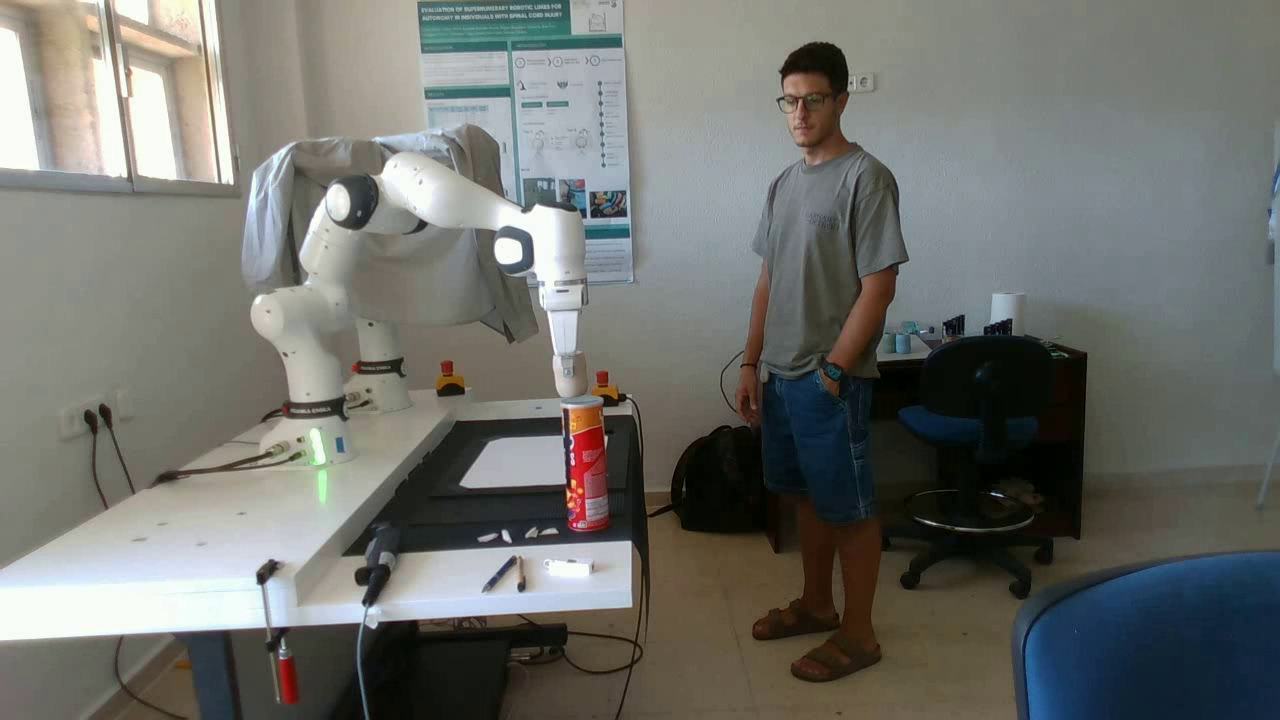}
    \caption{Representative examples from the recorded task, illustrating four distinct actions: robot picking up an object (top-left), human-robot handover (top-right), robot moving with a cup (bottom-left), and robot moving without a cup (bottom-right).}
    \label{fig:Dataset Example}
\end{figure}

Given the scarcity of publicly available datasets in the domain of human-robot interaction, we collected a custom dataset tailored to our target task. We designed a simple yet challenging Pick-and-Place scenario: a human hands a cup to the robot, which then grasps the cup and places it on top of another object. The robot subsequently retrieves the cup and returns it to the person, as shown in Fig. \ref{fig:Dataset Example}.

Data was collected from six participants, each performing the task twelve times, resulting in a total of 72 video recordings. Naturally occurring execution failures were intentionally preserved, providing realistic anomalous samples for later evaluation. These anomalies included events such as the robot dropping the cup, collisions with surrounding objects, and the unexpected presence of an additional person in the scene. In total, 17 such failure cases were recorded. 

During each trial, we simultaneously captured third-person view video and internal robot data via ROS bag files. From the robot, we specifically recorded joint torque values and gripper positions. Both video and sensor data were timestamped to ensure temporal alignment, enabling accurate multimodal analysis.

\begin{figure}
    \centering
    \includegraphics[width=0.49\linewidth]{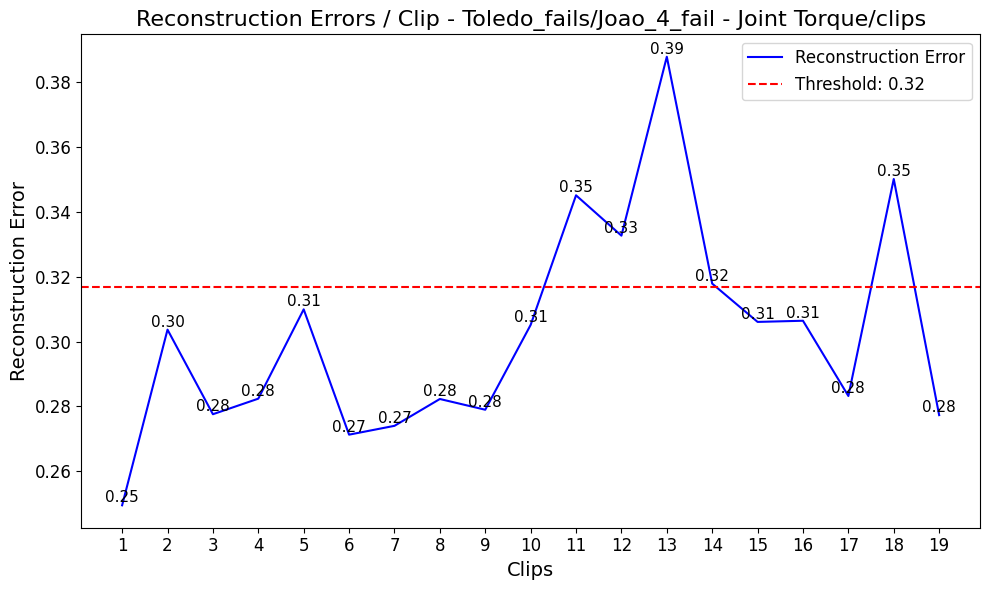}
    \includegraphics[width=0.49\linewidth]{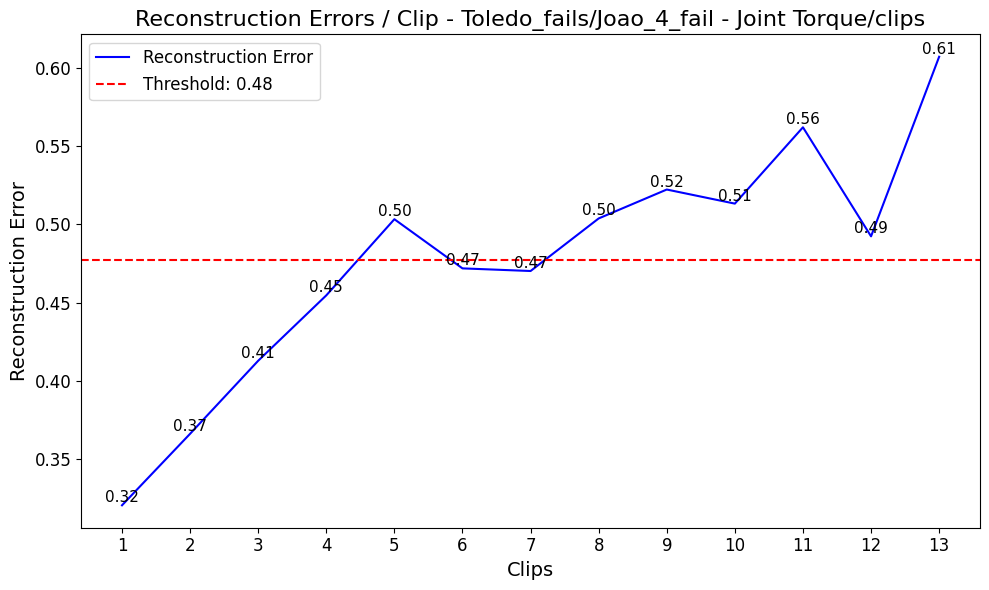}
    \caption{Reconstruction errors per clip during a joint torque failure. The left plot shows errors from the vision-only model, while the right plot shows errors from the multimodal model. The failure occurs at clip 11. The multimodal plot contains fewer clips due to robot shutdown following the internal failure.}
    \label{fig:Joint torque images}
\end{figure}

\begin{figure}
    \centering
    \includegraphics[width=0.49\linewidth]{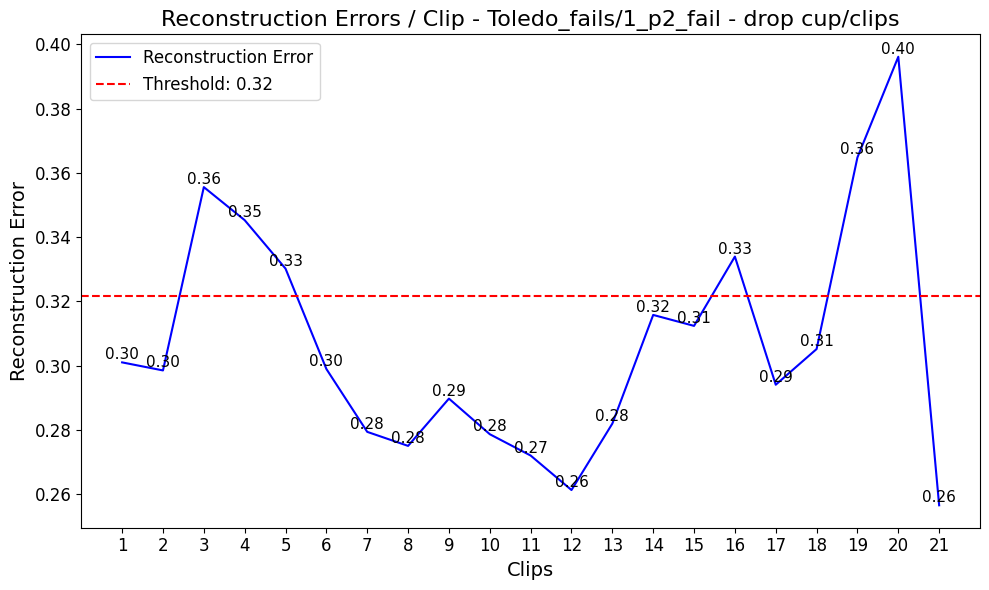}
    \includegraphics[width=0.49\linewidth]{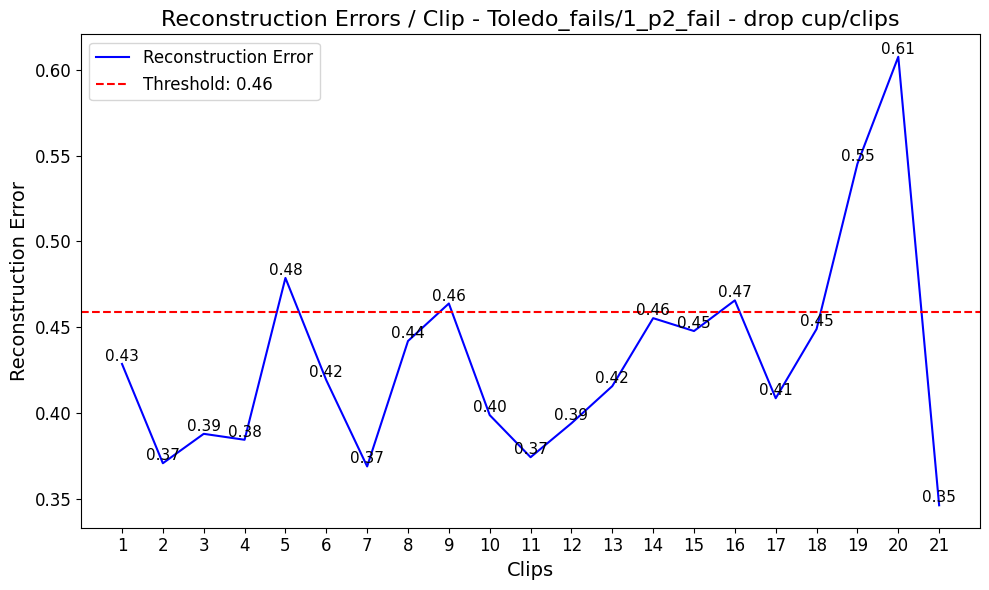}
    \caption{Reconstruction errors per clip during a failure event in which the robot drops the cup at clip 19. The left plot shows results from the vision-only model, while the right plot presents the multimodal model's reconstruction.}
    \label{fig:drop cup images}
\end{figure}
\subsection{Ablation Study}

For the ablation study, we evaluated the vision-only model, the vision+sensor model, vision+Scene Graph model and all modalities model on the set of videos containing anomalies. One video was excluded from the evaluation due to a missing ROS bag file.

\begin{figure}[!h]
    \centering
    \includegraphics[width=0.8\linewidth]{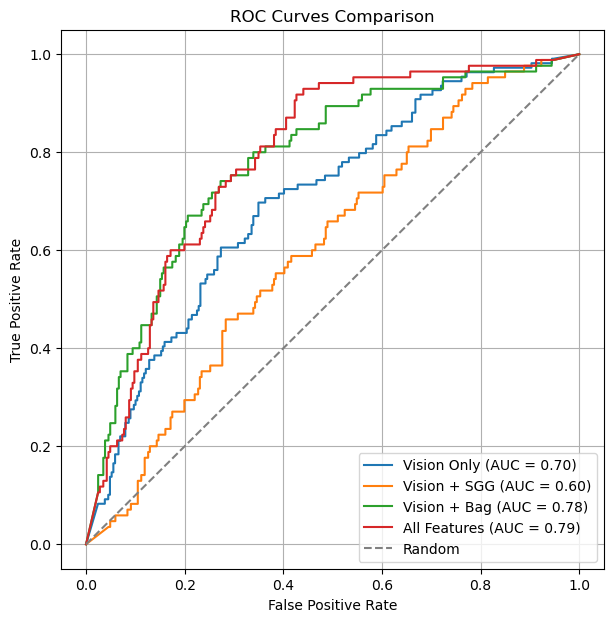}
    \caption{ROC Curves from the ablation study.}
    \label{fig:roc}
\end{figure}

For each model, we produced a ROC curve as shown in Fig.\ref{fig:roc}. From analysing the curves it is obvious that the model benefits specially from the addition of the sensor modality. Also, it is notable by the Vision + SGG curve, that the Scene Graph is actually punishing the models performance. We believe that is due to the low performance of the Scene Graph Generator and also the failures present in our dataset not being "relational".

Additionally, the Vision + Sensor model showed superior performance in cases where anomalies were closely tied to joint torque behaviour. For instance, in the scenario illustrated in Fig. \ref{fig:Joint torque images}, the anomaly stemmed from the robot reaching its joint limits. The multimodal model more clearly captured this prolonged stress through its reconstruction error pattern.

Furthermore, inspection of the reconstruction plots reveals that the inclusion of sensor data helps mitigate false spikes in the error signal, instances where the vision-only model momentarily misidentifies normal behaviour as anomalous. An example of this improved robustness can be seen in Fig. \ref{fig:drop cup images}, where the vision+sensor model maintains stability despite visual fluctuations.

\section{Conclusion}
Single-modality anomaly detection models are often limited in their ability to identify the full spectrum of anomalies. In this work, we demonstrate that incorporating additional data modalities significantly enhances a model’s ability to detect failures while reducing false positives. Our proposed framework, MADRI, serves as a proof of concept within the domain of Human-Robot Interaction, showing that combining visual information with robot sensor data yields more accurate and robust anomaly detection results. Future work will focus on expanding the dataset with more diverse tasks, integrating additional modalities, and extending the model to perform detailed failure classification.

\section*{Acknowledgment}

We would like to thank all the participants in the recording of the dataset and all the staff from the Hospital Nacional de Parapléjicos de España in Toledo.

\bibliography{bib}
\bibliographystyle{IEEEtran}

\end{document}